\documentclass[12pt]{elsarticle}
\usepackage{graphicx}
\usepackage{amssymb}
\usepackage{lineno}

\usepackage{amsmath,amssymb,amsfonts}
\usepackage{algorithmic}
\usepackage{graphicx}
\usepackage{textcomp}
\usepackage{xcolor}

\usepackage{algorithm}
\usepackage{algorithmic}
\usepackage{natbib}
\usepackage{multirow}

\usepackage[normalem]{ulem}
\useunder{\uline}{\ul}{}

\usepackage{tablefootnote}

\journal{Neurocomputing}

\begin{document}
	
%	\begin{frontmatter}
		\title{Imputation of Missing Data with Class Imbalance using Conditional Generative Adversarial Networks}

	\author{Saqib Ejaz Awan}
	\author{Mohammed Bennamoun}
	\author{Ferdous Sohel}
	\author{Frank M Sanfilippo}
	\author{Girish Dwivedi}
	%	\author[adr1]{Saqib Ejaz Awan \corref{cor1}}
	%	\ead{saqibejaz.awan@research.uwa.edu.au}
	%	\cortext[cor1]{Corresponding author}
	%	
	%	\author[adr1]{Mohammed Bennamoun}
	%	\ead{mohammed.bennamoun@uwa.edu.au}
		
	%	\author[adr2]{Ferdous Sohel}
	%	\ead{f.sohel@murdoch.edu.au}
		
	%	\author[adr3]{Frank Mario Sanfilippo}
	%	\ead{frank.sanfilippo@uwa.edu.au}
		
	%	\author[adr4,adr5,adr6]{Girish Dwivedi}
	%	\ead{girish.dwivedi@perkins.uwa.edu.au}
		
	%	\address[adr1]{Department of Computer Science and Software Engineering, The University of Western Australia, Crawley, WA 6009, Australia.}
	%	\address[adr2]{Discipline of Information Technology, Mathematics & Statistics, Murdoch University, Murdoch, WA 6150, Australia.}
	%	\address[adr3]{School of Population and Global Health, The University of Western Australia, Crawley, WA 6009, Australia.}
	%	\address[adr4]{Harry Perkins Institute of Medical Research, The University of Western Australia, Crawley, WA 6009, Australia.}
	%	\address[adr5]{Fiona Stanley Hospital, Murdoch, WA 6150, Australia.}
	%	\address[adr6]{Medical School, The University of Western Australia, Crawley, WA 6009, Australia.}
		\maketitle
		\begin{abstract}
			Missing data is a common problem faced with real-world datasets. Imputation is a widely used technique to estimate the missing data. State-of-the-art imputation approaches %, such as Generative Adversarial Imputation Nets (GAIN), 
			model the distribution of observed data to approximate the missing values. Such an approach usually models a single distribution for the entire dataset, which overlooks the class-specific characteristics of the data. Class-specific characteristics are especially useful when there is a class imbalance.
			We propose a new method for imputing missing data based on its class-specific characteristics by adapting the popular Conditional Generative Adversarial Networks (CGAN). Our Conditional Generative Adversarial Imputation Network (CGAIN) imputes the missing data using class-specific distributions, which can produce the best estimates for the missing values. We tested our approach on benchmark datasets and achieved superior performance compared with the state-of-the-art and popular imputation approaches. 
		\end{abstract}
		
		
%	\end{frontmatter}
	
	
	
	\section{Introduction}
	\label{sec: introduction}
	The growing use of machine learning and deep learning techniques demand more and more data. One big challenge associated with real-world data is missing values of certain attributes. The reasons of missingness in real-world data include an equipment failure, data corruption, privacy concerns of users, or a human error \cite{salgado2016missing, tran2017missing}. Missing data problem is categorised into three types based on the relationship between the missing and the observed values: Missing Completely at Random (MCAR), Missing at Random (MAR), and Missing Not at Random (MNAR) \cite{kwak2017statistical, mesquita2017euclidean}. MCAR occurs when the missingness is totally independent of \emph{all the variables} present in the data. In MAR, the missingness is related to only the \emph{observed variables}. MNAR exists when the missingness is dependent on both the \emph{observed and missing variables}.
	
	To perform a task, e.g. classification or prediction, statistical and machine learning algorithms generally require complete data \cite{salgado2016missing, ruiz2018machine, smieja2018processing}. It highlights the need to handle missing data properly. A simple approach to achieve completeness is the \textit{complete-case analysis}, which only uses the observed (non-missing) values of the data \cite{jakobsen2017and}. This approach is suitable if a few samples of data contain missing values and produces biased results otherwise \cite{jakobsen2017and}. Another approach to address the missing data is to replace it with plausible approximations learnt from the observed data. This approach is called \emph{missing data imputation} \cite{salgado2016missing}. Simple imputation approaches replace the missing values of a variable/column with a statistical estimate such as mean or median of all the non-missing values of that variable/column. These approaches replace all the missing values in a variable with the \textit{same estimated value} thus underestimating the variance of the imputed values leading to poor performance. Advanced approaches, such as Multiple Imputation by Chained Equations (MICE) \cite{buuren2010mice}, explore the correlation between the variables to better approximate the missing values. Some joint modelling approaches, such as Expectation Maximization (EM), assume a multivariate normal distribution and assert a joint distribution on the entire data to impute missing values \cite{rahman2016missing}. 
	
	Another naturally inherent problem of the real-world data is the skewed class distributions. It is a condition, commonly known as the \textit{class imbalance problem},  in which the majority of data belongs to one class and a significantly small amount of data belongs to the remaining classes \cite{khan2017cost}. For example, in a binary classification problem, it is naturally expected to have only a few positive cases of fraudulent transactions and a significantly large number of non-fraudulent transactions. Since most of the machine learning models are designed with the assumption of equal number of samples of each class, they over-classify the majority class and ignore the minority class \cite{nguyen2009learning}. In most of the cases, the minority class in a real-world data is the class of interest \cite{napierala2016types} e.g., detecting a fraudulent transaction or a cancerous image. Thus, the performance of these analytical models degrade as the class imbalance problem grows in the real-world data.

	Imputation of missing data in imbalanced datasets is a challenging task because only a few samples are available from the minority class. In this case, an advanced imputation technique, such as Generative Adversarial Imputation Network (GAIN) \cite{yoon2018gain}, models one distribution for the entire data. However, the characteristics of one class samples may differ from the characteristics of another class samples. Such an approach will not produce high performance (Section \ref{subsec:CGAINvsGAIN}). This issue is further exacerbated if the data has class imbalance i.e., one class has more samples than the others. Therefore, intuitively, an approach which takes into account the individual class distributions can potentially provide better imputation performance. This leads us to propose our Conditional Generative Adversarial Imputation Network (CGAIN) approach which aims to impute the missing data conditional to its class, making the imputation process rely on the individual class characteristics for missing data imputation. The main contributions of this paper are: 
	\begin{itemize}
		\item A novel missing data imputation technique that incorporates the class distribution of the missing data.
		\item State-of-the-art imputation performance on imbalanced datasets.
	\end{itemize}
	
	The rest of this paper is structured as follows. We briefly discuss the related work on missing data imputation in Section \ref{sec: related work}. We introduce our proposed approach in Section \ref{sec: proposed approach} and present the experimental results and analysis in Section \ref{sec: experiments}. Section \ref{sec: conclusion} concludes this work.
	
	\section{Related work}
	\label{sec: related work}
	Imputation of missing data is an active research area. Several imputation approaches have been proposed to complete the missing data for various applications such as medical, image inpainting, and financial data \cite{ivanov2018variational,chen2018graph,lin2020missing}. Missing data imputation techniques are commonly divided into two groups: \textit{single imputation} and \textit{multiple imputation}. Single imputation replaces the missing values with estimated values only once, while multiple imputation repeats the imputation process several times and combines the results of all imputations in the end. 
	
	\underline{Single imputation approaches} are further classified into univariate single imputation and multivariate single imputation approaches. Univariate approaches use the observed values of the same column to replace the missing values with a statistical estimate such as mean, median, the most frequent value, and the last observation carried forward \cite{zhang2016missing}. This process, however, undermines the variance of the imputed values. Multivariate imputation approaches use the correlation between different columns of data and use it to impute the missing values. A common approach is to predict the missing values in a column using a regression model based on the values in the observed columns. This process is then repeated for all the columns with the missing values to complete the data.
	
	\underline{Multiple imputation} allows for the uncertainty in the missing data by creating multiple different plausible imputed data sets and combining the results obtained from each of them at the end \cite{sovilj2016extreme}. First, multiple copies of the data are created each containing missing data replaced with imputed values.Then, analytical methods are used to fit the model of interest to each of the imputed datasets and the final results are produced by combining the results from all the copies of data.
	
	With the recent advancements in machine learning and deep learning, new imputation approaches have been proposed which improve the imputation performance of the existing approaches. These approaches have limited the use of statistical based imputation approaches. However, new approaches continue to emerge which claim to surpass the performance of existing imputation methods.
	Popular new imputation approaches can broadly be categorized into
	\textit{discriminative} and \textit{generative} methods \cite{Tarap2019HitAM}. The discriminative methods learn the boundary between classes of data such as support vector machines and decision trees. The generative methods focus on how the data is actually generated to model the distribution of data.
	The discriminative methods model the decision boundary between different classes present in data to perform imputation. These approaches include \emph{MissForest} and \emph{Matrix Completion} \cite{mazumder2010spectral,troyanskaya2001missing, cai2010singular, candes2009exact, mitra2010large}.
	The generative methods model the actual distribution of data to perform imputation. Popular generative imputation approaches include MICE, denoising autoencoders, and Gaussian mixture models \cite{buuren2010mice, lu2020multiple, sovilj2016extreme}.
	
	Recently, GAIN \cite{yoon2018gain} has been introduced for imputation, which combines both the generative and discriminative models in an adversarial manner. The generative and discriminative models in this approach compete with each other to excel their tasks \cite{yoon2018gain}. The aim of this method is to achieve a generative model, which can produce new samples whose distribution is so close to the original data distribution that the discriminative model is unable to distinguish between a real sample and a sample generated by the generative model. The generator in GAIN receives two inputs: input data and mask. A mask represents the presence/absence of a value. A presence is marked as 1, while an absence is 0. The discriminator's output predicts the complete mask whose elements show the possibility of the corresponding input value to be observed, unlike a standard discriminator of Generative Adversarial Network (GAN), which tells only if the input to discriminator is real or fake as a whole. Their approach introduces a hint mechanism, which becomes an input of the discriminator. This approach learns to model the distribution of entire data as a whole which may overlook the unique characteristics of a minority class in the case of an imbalanced data.
	
	In this work, we propose a new imputation approach which aims to learn the unique class-specific characteristics and use them to impute missing data from that class. The data imputed using our approach will be based on individual class distributions rather than the entire data distribution and we hypothesize that this will produce more accurate estimates of the missing data.

	\section{Proposed approach: CGAIN}
	\label{sec: proposed approach}
	
\begin{figure}[htbp]
		% \centerline{\includegraphics[width=80mm]{block_diagram.png}}
		\centerline{\includegraphics[width=70mm]{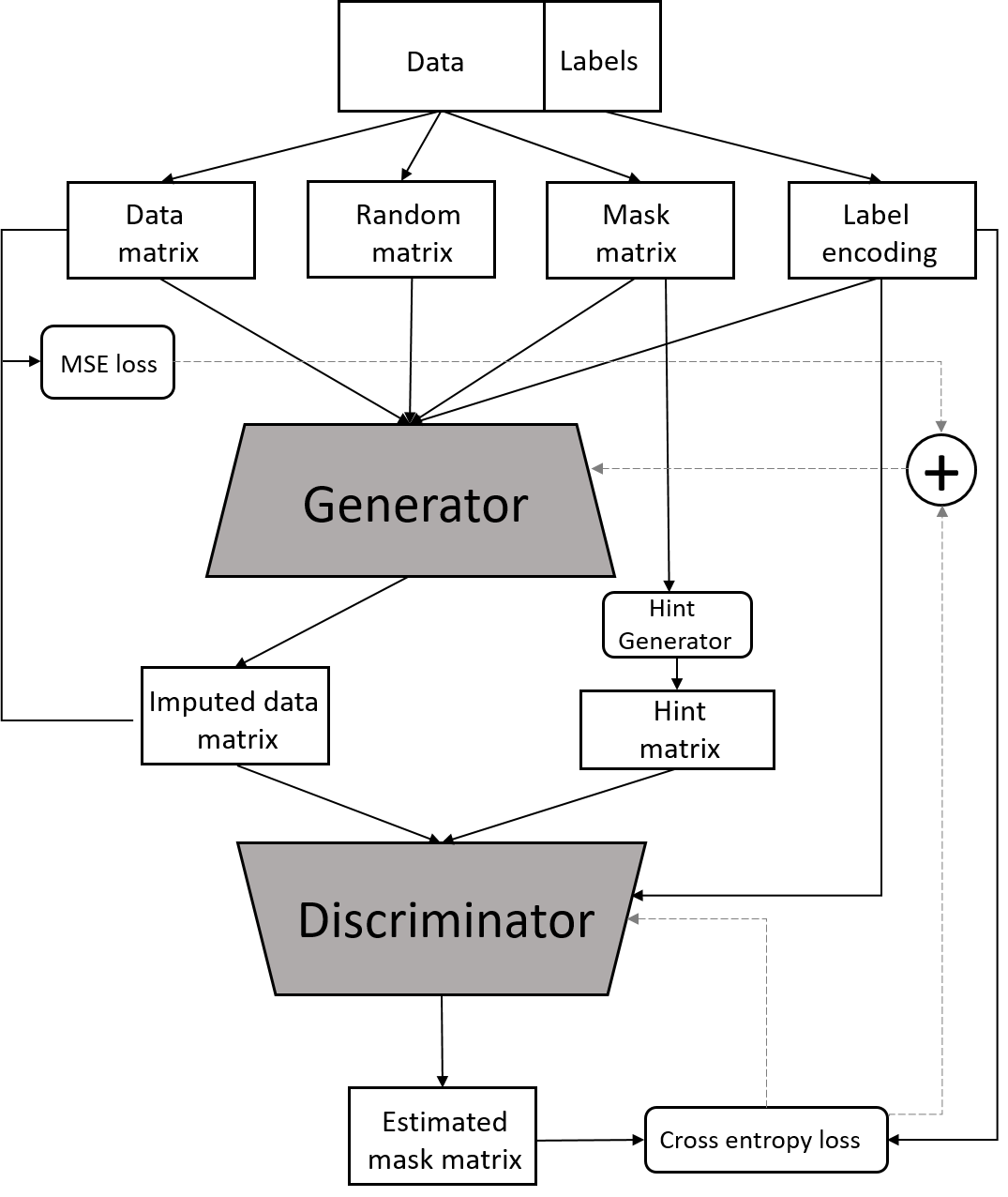}}
		\caption{Block diagram of the proposed CGAIN approach.}
		\label{fig}
	\end{figure}

	Our proposed CGAIN approach aims at producing a generator which can produce fake data pertaining to a class. This means that our generator can not only produce data but it is also aware of which class of data it has to produce. We train the generator in an adversarial manner with a discriminator. The discriminator receives the fake data and predicts, for each data element, whether it was missing or observed in the original data. This step forces the generator to produce fake values, against the missing values of data, which are very close to the original data distribution. Figure \ref{fig} shows the block diagram of our approach. The generator produces fake data using the original data with missing values, class labels, and random data. This fake data contains values against both the missing values and observed values in the original data. The discriminator receives this data and predicts which components were originally missing. The generator is given the feedback (using cross-entropy loss) on how successful the discriminator was in discriminating missing values from observed values. The generator also receives the mean squared error of the fake and original values. Based on these feedbacks, the generator adjusts its parameters and attempts to produce fake samples which appear as real samples to fool the discriminator. At the end of the training phase, we get a generator which is capable of producing realistic (fake) values against the missing values in the original data.
	
	\subsection{Problem Formulation for Missing Data Imputation}
	Suppose we represent our data with a \emph{data vector} $ \textbf{X} = (\textit{X}_1, \textit{X}_2, \ldots , \textit{X}_d) $, where $\textbf{X}$ is a random variable in a \emph{d}-dimensional \emph{data space} $\mathcal{X} = \mathcal{X}_1 \times \mathcal{X}_2 \times \ldots \times \mathcal{X}_d $ and \emph{d} denotes the total number of data samples. 
	We represent the \emph{labels/outcomes} of our data as a \emph{label vector} $\textbf{Y} = (Y_1, Y_2, \ldots, Y_d)^T$, where each vector element takes the form $ \{0,1\}^m $ and $m$ is the total number of classes/outcomes. 
	Also assume a \emph{mask vector} $\textbf{M}=(M_1, M_2, \ldots , M_d)$,  which takes its values from $\{0,1\}^d$. 
	We can define a \emph{missing data vector} $ \Tilde{\textbf{X}} = (\Tilde{\textit{X}}_1, \Tilde{\textit{X}}_2, \ldots , \Tilde{\textit{X}}_d) $, %in a \emph{missing data space} $ \Tilde{\mathcal{X}} = \Tilde{\mathcal{X}}_1 \times \Tilde{\mathcal{X}}_2 \times  \ldots \times \Tilde{\mathcal{X}}_d $ %
	which replaces each missing value with an asterisk $(*)$ as shown in Equation \ref{eqn: tilde X}.
	
	% \hline
	
	% Suppose we have, \textbf{1)}  a \emph{data vector}  $ \textbf{X} = (\textit{X}_1, \textit{X}_2, \ldots , \textit{X}_d) $  as a random variable %whose distribution is denoted as $P(\textbf{X})$. $\textbf{X}$% 
	% which takes its values from a \textit{d}-dimensional \emph{data space} $\mathcal{X} = \mathcal{X}_1 \times \mathcal{X}_2 \times \ldots \times \mathcal{X}_d $. \textbf{2)}   a \emph{class/outcome vector}  represented as $\textbf{Y} = (Y_1, Y_2, \ldots, Y_d)^T$ which refers to the class/outcome labels of the form $ \{0,1\}^m $, where $m$ is the total number of classes/outcomes. \textbf{3)} a random variable, named \emph{mask}, $\textbf{M}=(M_1, M_2, \ldots , M_d)$  which takes its values from $\{0,1\}^d$.
	% Lets define a \emph{ missing data vector} $ \Tilde{\textbf{X}} = (\Tilde{\textit{X}}_1, \Tilde{\textit{X}}_2, \ldots , \Tilde{\textit{X}}_d) $, in a \emph{missing data space} $ \Tilde{\mathcal{X}} = \Tilde{\mathcal{X}}_1 \times \Tilde{\mathcal{X}}_2 \times  \ldots \times \Tilde{\mathcal{X}}_d $, such as:
	
	\begin{equation}
	\label{eqn: tilde X}
	\Tilde{\textit{X}}_i = 
	\left\{ 
	\begin{array}{ll}
	\textit{X}_i, &  {\rm if} \, \textit{M}_i= 1\\
	*, & {\rm otherwise} \\
	\end{array} 
	\right.
	\end{equation}
	
	% where $*$ represents a missing value, a point not present in any $ \Tilde{\mathcal{X}}_i $. 
	The goal of an imputation approach is to \emph{impute} the missing values in %$\Tilde{x}_i$ (where $\Tilde{x}_i$ are the realizations of the random variable%
	$\Tilde{\textbf{X}}$. Our missing data imputation aims at generating samples according to the conditional probability of $\textbf{X}$ given $\Tilde{\textbf{X}}$ and \textbf{Y} i.e., $P(\textbf{X} | \Tilde{\textbf{X}}, \textbf{Y} )$.

	%\subsection{Imputation of missing data}
	
	% In an imputation setting, our goal is to impute the missing values in each $x_i$.
	
	\subsection{Proposed Conditional Generative Adversarial Imputation Networks}
	This section describes our CGAIN based approach to simulate the $P(\textbf{X} | \Tilde{\textbf{X}}, \textbf{Y} )$. Figure \ref{fig} shows a concise summary of our approach. Our approach contains one generator and one discriminator. Our generator generates fake data using the original data with missing values, a mask which locates the missing values, the encoded class-labels, and a noise matrix. The discriminator discriminates between the observed and the missing values in the data by predicting the mask matrix. The main components of our approach are discussed below.
	
	\subsubsection{Generator ($G$)}
	The Generator receives the input data with missing values $\Tilde{\textbf{X}}$, mask \textbf{M}, labels \textbf{Y}, and noise \textbf{Z} to output a vector of imputations $\Bar{{\textbf{X}}}$. The \emph{noise vector} $\textbf{Z} = (Z_1, Z_2, \ldots , Z_d)$ contains \emph{d}-dimensional noise while \textbf{Y} is a one-hot encoded vector. The generator $G$ can be mathematically defined as %$G: \Tilde{\mathcal{X}} \times \{0,1\}^d \times [0,1]^d \times \{0,1\}^m \to \mathcal{X} $.% 
	$G: \Tilde{\mathcal{X}} \times \mathcal{M} \times \mathcal{Z} \times \mathcal{Y} \to \mathcal{X} $, where $\mathcal{M}$, $\mathcal{Z}$, and $\mathcal{Y}$ denote the distributions of mask, noise, and labels/outcomes, respectively.
	
	Since G outputs a value for every component rather than producing an estimate of the missing values only, we can create a \textit{completed data vector} $\hat{\textbf{X}}$, which combines the observed values from $\Tilde{\textbf{X}}$ and the imputed values from $\Bar{\textbf{X}}$ as given in Equations \ref{eqn: G output} and \ref{eqn: G data}. 
	
	\begin{equation}
	\label{eqn: G output}
	\Bar{\textbf{X}} = G(\Tilde{\textbf{X}}|\textbf{Y},\textbf{M},(\textbf{1}-\textbf{M}) \odot \textbf{Z})
	\end{equation}
	\begin{equation}
	\label{eqn: G data}
	\hat{\textbf{X}} = \textbf{M} \odot \Tilde{\textbf{X}} + (\textbf{1}-\textbf{M}) \odot \Bar{\textbf{X}}
	\end{equation}
	
	where $\odot$ represents element-wise multiplication and $\textbf{1}$ denotes a \textit{d}-dimensional vector of 1s. As seen in Equation \ref{eqn: G data}, $\hat{\textbf{X}}$ takes the observed values from $\Tilde{\textbf{X}}$ and replaces each $*$ with its corresponding value from $\Bar{\textbf{X}}$. This setup is inspired by a standard GAN and the generator used by \cite{yoon2018gain}.
	
	\subsubsection{Discriminator ($D$)}
	The discriminator is generally introduced as an adversary to train the generator. Conventionally, there is only one output of $D$ i.e., either \textit{completely real} or \textit{completely fake}. However, the output of $D$ in an imputation setting contains multiple components, some of which are real while others are fake. So, our $D$ tries to distinguish the real (observed) components from the fake (missing) components. This is achieved by predicting a mask vector \textbf{m}. We can then compare this predicted mask with the original mask \textbf{M}. Formally, our $D$ can be mathematically defined as $D: \mathcal{X} \times \mathcal{Y} \to [0,1]^d$, where $[0,1]^d$ represents the predicted mask vector \textbf{m}.%where the \emph{i}-th component of \textit{D} corresponds to its probability of being observed/not missing conditional on the class label.
	
	\subsubsection{Hint ($H$)}
	We also use a hint mechanism similar to Yoon et al. \cite{yoon2018gain} in our approach. This hint is expressed as a random variable \textbf{H} which takes its values in a \emph{hint space} $\mathcal{H}$. The hint vector supports \emph{D} by telling it some of the imputed and observed values which allows the \emph{D} to decide whether other values are imputed or observed. \textbf{H} is passed as an additional input to \emph{D}  which is then mathematically expressed as $D: \mathcal{X} \times \mathcal{H} \times \mathcal{Y} \to [0,1]^d$.
	The hint is deemed necessary since \emph{G} can produce several distributions and for all of those \emph{D} can not distinguish between a real and a fake value. Therefore, giving a hint \textbf{H} to \emph{D} restricts the solution to a single distribution. \textbf{H} is obtained using Equation \ref{eqn: H}.
	
	\begin{equation}
	\label{eqn: H}
	\textbf{H} = \textbf{B} \odot \textbf{M} + 0.5 \odot (1 - \textbf{B})
	\end{equation}
	
	where $B \in \{0,1\}^d$ is a random variable obtained by uniformly sampling \emph{k} from $\{1,2, \ldots, d\}$ and applying Equation \ref{eqn: B}. The term $0.5$ in Equation \ref{eqn: H} represents a hint value similar to that used by Yoon et al. \cite{yoon2018gain}.
	
	\begin{equation}
	\label{eqn: B}
	\textbf{B}_j = 
	\left\{ 
	\begin{array}{ll}
	1, &  {\rm if} \, j \neq k\\
	0, & {\rm if} \, j = k\\
	\end{array} 
	\right.
	\end{equation}
	
	\subsubsection{The objective function}
	The objective function of our CGAIN approach has two parts as inspired by the standard Conditional Generative Adversarial Network (CGAN) \cite{mirza2014conditional}. First, we train \emph{D} to maximize the correct prediction of \textbf{M}. Secondly, we train \emph{G} to minimize the probability of \emph{D} correctly predicting \textbf{M}. The overall objective function and loss function of our CGAIN approach are given in Equations \ref{eqn: obj function brief} and \ref{eqn: obj function}, respectively.
	
	\begin{equation}
	\label{eqn: obj function brief}
	\displaystyle{\min_{G} \max_{D}} \; \mathcal{L}(D,G)
	\end{equation}
	
	\begin{equation}
	\label{eqn: obj function}
	\mathcal{L}(D,G) = \mathbb{E}_{\hat{\textbf{X}},\textbf{M},\textbf{H},\textbf{Y}} \left[\textbf{M}^T \; log \; D  ((\hat{\textbf{X}},\textbf{H})|\textbf{Y}) + (\textbf{1}-\textbf{M})^T \; log \;(1-D (\hat{\textbf{X}},\textbf{H})|\textbf{Y}))\right]
	\end{equation}
	
	Since the output of \emph{D} can be expressed as $\hat{\textbf{M}}=\emph{D}((\hat{\textbf{X}},\textbf{H})|\textbf{Y})$, the loss function of \emph{D} can be expressed by the cross entropy Equation \ref{eqn: loss D}. 
	
	\begin{equation}
	\label{eqn: loss D}
	\mathcal{L}_D = \sum_{\forall i:b_i=0} \large[ m_i \;  log(\hat{m}_i) + (1-m_i) \; log(1-\hat{m}_i) \large]
	\end{equation}
	where the term $b_i=0$ corresponds to those values of $\hat{\textbf{M}}$ for which $\textbf{H}$ is 0.5 according to Equation \ref{eqn: H}. This ensures \emph{D} to learn those mask values for which the absolute hint value (0 for missing, 1 for not missing) was not provided. %\emph{D} is trained according to the objective function given in Equation \ref{eqn: train D}.
	
	% \begin{equation}
	% \label{eqn: train D}
	%     \min_{D} - \sum_{j=1}^{k_D} \mathcal{L}_D (m(j), \hat{m}(j), b(j))
	% \end{equation}
	
	% recalling that $\hat{m}(j)=D(\hat{x}(j),m(j))$
	
	Similar to \cite{yoon2018gain}, the loss function of \emph{G} comprises of two parts since the output of \emph{G} contains imputed values for both the missing values and the observed values. The first part is the loss of imputed values %($\mathcal{L}_{imp}$) 
	whereas the second part is the loss of the observed values %($\mathcal{L}_{obs}$). 
	The combined loss function $\mathcal{L}_G$ is given in Equation \ref{eqn: loss comb}.
	
	% \begin{equation}
	% \label{eqn: loss Gimp}
	%     \mathcal{L}_{imp}=\sum_{\forall i:b_i=0} (1-m_i) \; log(\hat{m}_i)
	% \end{equation}
	
	% \begin{equation}
	% \label{eqn: loss Gobs}
	%     \mathcal{L}_{obs}=\sum_{i=1}^{d} m_i \; L_{obs}(x_i,x_i^\prime) \large]
	% \end{equation}

	% The combined loss of \emph{G} is then given as a weighted sum of $\mathcal{L}_{imp}$ and $\mathcal{L}_{obs}$ as shown in Equation \ref{eqn: loss comb}, 
	
	\begin{equation}
	\label{eqn: loss comb}
	\mathcal{L}_G= \sum_{\forall i:b_i=0} (1-m_i) \; log(\hat{m}_i) \; + \alpha \; \sum_{j=1}^{d} m_i \; L_{obs}(x_i,x_i^\prime)
	\end{equation}
	
	where, similar to Yoon et al. \cite{yoon2018gain}, $\alpha$ is a positive hyper-parameter and $L_{obs}(x_i,x_i^\prime)$ is given in Equation \ref{eqn: loss Gobs2}:
	
	\begin{equation}
	\label{eqn: loss Gobs2}
	L_{obs}(x_i,x_i^\prime) = 
	\left\{ 
	\begin{array}{ll}
	(x_i^\prime-x_i)^2 &  {\rm if} \, x_i \; is \; continuous\\
	-x_i \; log(x_i^\prime) & {\rm if} \, x_i \; is \; binary\\
	\end{array} 
	\right.
	\end{equation}
	
	\subsection{CGAIN Algorithm}
	The training of our proposed CGAIN algorithm is inspired by the original GAN approach, which iteratively trains \emph{D} and \emph{G}. We designed \emph{G} and \emph{D} as fully connected neural networks with two hidden layers. We kept the number of neurons in each hidden layer as three times the number of columns/features in the input data. 
	
	We first optimized \emph{D} with a fixed \emph{G} using mini-batches of (128 samples of) data. For every mini-batch including the corresponding labels \textbf{Y}, \emph{n} independent samples of \textbf{Z}, \textbf{B}, and \textbf{M} are drawn to compute the imputed data $\Tilde{\textbf{X}}$ according to Equations \ref{eqn: G output} and \ref{eqn: G data}. Then hint vector \textbf{H} is produced using Equation \ref{eqn: H}. Then, the estimated mask $\hat{\textbf{M}}$ is obtained using $\emph{D}((\hat{\textbf{X}},\textbf{H})|\textbf{Y})$ followed by the optimization of \emph{D}. 
	
	The next step is to update \emph{G} by keeping the newly trained \emph{D} fixed. Again, \emph{n} independent samples of \textbf{Z}, \textbf{B}, and \textbf{M} are drawn for every mini-batch to compute \textbf{H} and update \emph{G}. The CGAIN algorithm is presented in Algorithm 1.
	% \ref{algo}
	
	% \begin{algorithm}[H]
	% \caption{Pseudo-code of CGAIN}
	% \begin{algorithmic}
	%     \While{training loss has not converged}
	%     {
	%     (1) Discriminator optimization \\
	%     (2) Generator optimization
	%     }
	% \end{algorithmic}
	% \end{algorithm}
	
	\begin{algorithm}[ht]
		%\SetAlgoLined
		
		\caption{Pseudo-code of the proposed CGAIN algorithm}
		\hspace*{\algorithmicindent} \textbf{Input:} Discriminator batch size $n_D$, Generator batch size $n_G$\\
		\hspace*{\algorithmicindent} \textbf{Output: }Trained CGAIN algorithm 
		
		\begin{algorithmic}
			\WHILE{training loss does not converge}
			\STATE \textbf{(A) Discriminator optimization} \;
			\STATE Draw $n_{D}$ data samples from dataset $\{(\Tilde{x},y,m)\}$, noise samples $z$ from $\textbf{Z} $, hint samples $b$ from $\textbf{B} $     \;
			% Draw $n_{D}$ noise samples $z$ from $\textbf{Z} $           \newline
			% Draw $n_{D}$ hint samples $b$ from $\textbf{B} $            \newline
			\FOR{$i=1,\ldots, n_{D}$}
			\STATE $x_i \leftarrow G(\Tilde{x}_i, y_i, m_i, z_i)$
			\STATE $\Hat{x}_i \leftarrow m_i \odot \Tilde{x}_i + (1-m_i) \odot \Bar{x}_i$
			\STATE $h_i \leftarrow b_i \odot m_i + 0.5(1-b_i)$
			\ENDFOR
			\STATE Update \emph{D} using stochastic gradient descent (SGD) \newline
			% $\mathcal{L}_\emph{D}(m(j), \emph{D}(\hat{x}(j), h(j), b(j))$
			
			\STATE \textbf{(B) Generator optimization}
			\STATE Draw $n_{D}$ data samples from dataset $\{(\Tilde{x},y,m)\}$, noise samples $z$ from $\textbf{Z} $, hint samples $b$ from $\textbf{B} $     %\newline
			\FOR{$i=1,\ldots, n_{G}$}
			\STATE $h_i = b_i \odot m_i + 0.5(1-b_i)$
			\ENDFOR 
			\STATE Update \emph{G} using stochastic gradient descent (SGD) with fixed \emph{D} %\newline
			% $\mathcal{L}_\emph{G}(m(j), \hat{m}(j),b(j)) + \alpha \mathcal{L}_M (x(j), \Tilde{x}(j))$
			\ENDWHILE
		\end{algorithmic}
	\end{algorithm}

	\section{Experiments and Results}
	\label{sec: experiments}
	We tested our proposed CGAIN approach with multiple publicly available real-world datasets available at the University of California Irvine (UCI) Machine Learning repository \cite{Dua:2019}. We compared our approach with the state-of-the-art GAIN approach \cite{yoon2018gain} and other popular imputation approaches. We also evaluated our approach on various percentages of missing data ranging from 5\% to 20\%. The missing data was created, in an MCAR style, by randomly removing values in all experiments.
	
	\subsection{Datasets and Methods}
	The details of the datasets used in this work are shown in Table \ref{tab: datasets}. We tested our approach on 5 UCI repository datasets. The choice of these datasets is mainly inspired by the experiments of Yoon et al. \cite{yoon2018gain}. This selection allows us to present a comparative analysis of the performance of our approach with the state-of-the-art GAIN approach. 
	
	The \emph{Breast Cancer} dataset contains features of  digitized cancerous images such as the radius of cell nuclei, texture, and perimeter etc. The \emph{Spambase} dataset contains email features such as the occurrence of a specific word in an email, length of sequences of consecutive capital letters etc. \emph{Letter recognition} dataset contains features from images of capital alphabets. \emph{Default credit card} dataset is a classification dataset for the prediction of default of a customer based on age, amount of given credit, and history of past payments etc. \emph{News popularity} dataset contains features of online news articles such as the number of words in its title, number of hyperlinks in the article, the average length of words etc.
	
	The Breast cancer, Spambase, Default credit card and News popularity datasets are binary datasets (two classes only). The letter recognition dataset is multi-class having 26 classes. Our results show improved imputation performance on binary as well as multi-class datasets highlighting the efficacy of our approach. 
	
	\begin{table}[htbp]
		% \centering
		\caption{Characteristics of datasets used in this work.\strut}
		\label{tab: datasets}
		\resizebox{\textwidth}{!}{%
			\begin{tabular}{|c|c|c|c|c|}
				\hline
				\textbf{Dataset} & \textbf{Number of instances} & \textbf{Number of classes} & \textbf{Majority class vs minority class (\%)}& \textbf{Number of attributes} \\ \hline
				Breast cancer       & 569                         &2& 62.74 vs 37.26 & 30                            \\
				Spambase            & 4,601                       &2& 60.60 vs 39.40 & 57                            \\
				Default credit card & 30,000                      &2& 77.88 vs 22.12 & 24                            \\
				News popularity     & 39,644                      &2& 53.36 vs 46.64 & 61                            \\ 
				% Diabetes            & 768                         & & 2& 8                            \\
				Letter recognition  & 20,000                      &26 & multi-class & 17                            \\
				\hline
			\end{tabular}
		}
	\end{table}
	
	We selected GAIN \cite{yoon2018gain}, MICE \cite{buuren2010mice}, MissForest \cite{stekhoven2012missforest}, and Matrix completion \cite{cai2010singular} approaches to compare with our proposed CGAIN approach. MICE is a popular statistical imputation approach, whereas GAIN, MissForest, and Matrix completion were the best-performing methods, on the datasets used in this work, in a recent study \cite{yoon2018gain}.
	
	\subsection{Performance of our proposed CGAIN}
	\label{subsec:CGAINvsGAIN}
	The comparative performance of our proposed CGAIN approach is given in Table \ref{tab:5 perc} to Table \ref{tab:20 perc}. We report the average Root Mean Squared Error (RMSE) of 10-fold cross-validated experiments. We compare our proposed CGAIN approach with the state-of-the-art GAIN approach \cite{yoon2018gain} and other popular imputation approaches. Table \ref{tab:5 perc} to Table \ref{tab:20 perc} show the RMSE of all the approaches where the proportion of missing data ranges from 5\% to 20\%. Our proposed CGAIN\footnote{https://github.com/saqibejaz/CGAIN.git} approach provided superior performance compared to other approaches on all the datasets. %The second best performance is underlined in the tables. 
	We used the publicly available GitHub\footnote{\label{gain_footnote}https://github.com/jsyoon0823/GAIN} code of GAIN in our experiments. Other approaches such as MICE, MissForest, and imputation using matrix completion were implemented using the publicly available python libraries (\emph{missingpy, sklearn, and matrix\_completion}). 
	
	Our CGAIN consistently outperforms all techniques. CGAIN provides a lower RMSE (mean ± std) of 0.0643 ± 0.0014, 0.0628 ± 0.0024, 0.0673 ± 0.0039, and 0.0637 ± 0.0092 compared with the second best RMSE of 0.0658 ± 0.0022, 0.0692 ± 0.0017, 0.0689 ± 0.0058, and 0.0726 ± 0.0038 at 5\%, 10\%, 15\%, and 20\% missing values of the \emph{Breast Cancer} dataset, respectively. For the \emph{Default Credit} dataset, our proposed CGAIN shows lower RMSE of 0.2329 ± 0.0039, 0.2009 ± 0.0022, 0.2314 ± 0.0035, and 0.2213 ± 0.0099 compared with the GAIN's 0.2428 ± 0.0093, 0.2109 ± 0.0344, 0.2442 ± 0.0089, and 0.2426 ± 0.0090 at 5\%, 10\%, 15\%, and 20\% missing values, respectively. 
	
	% The only abnormality to this trend is observed in the \emph{Spambase} dataset where our proposed approach yields a slightly greater RMSE of 0.0827 compared to GAIN's 0.0629. However, when a small percentage of data is missing, simpler approaches such as complete-case analysis work well. Missing data problem becomes more challenging when a large proportion of data is missing and that is where the performance of conventional imputation approaches deteriorate. Our approach, however, gets better as the proportion of missing data increases. For example, our CGAIN approach gives a comparable RMSE of 0.0855 compared with GAIN's 0.0803 at 10\% missing values and a superior performance with RMSE 0.0832 and 0.0904 compared with GAIN's RMSE of 0.0847 and 0.0915 at 15\% and 20\% missing values of the \emph{Spambase} dataset, respectively. This shows that our approach works well as the missing data problem becomes more severe.

	% \usepackage{graphicx}
	% \usepackage[normalem]{ulem}
	\begin{table}[htbp]
		\caption{RMSE performance (mean±std ) of our proposed CGAIN approach on 5\% missing data.\strut %\tablefootnote{Best results are shown in boldface, while the 2nd best results are underlined.}
		}
		\label{tab:5 perc}
		\resizebox{\textwidth}{!}{%
			\begin{tabular}{llllll}
				\hline
				\textbf{Dataset}        & \textbf{CGAIN}         & \textbf{GAIN}       & \textbf{MICE}       & \textbf{MissForest} & \textbf{Matrix} \\ \hline
				\textbf{Breast Cancer}  & \textbf{0.0643±0.0014} & 0.1372±0.0013       & 0.0854±0.0013       & {\ul 0.0658±0.0022} & 0.6881±0.0034   \\
				\textbf{Spambase}       & \textbf{0.0611±0.0060} & {\ul 0.0723±0.0018} & 0.0747±0.0045       & 0.0771±0.0071       & 0.0943±0.0009   \\
				\textbf{Letter}         & \textbf{0.1066±0.0078} & 0.1437±0.0029       & 0.1833±0.0008       & {\ul 0.1189±0.0018} & 0.4545±0.0020   \\
				\textbf{Default Credit} & \textbf{0.2329±0.0039} & {\ul 0.2428±0.0093} & 0.2479±0.0079       & 0.2902±0.0010       & 0.2565±0.0089   \\
				\textbf{News}           & \textbf{0.1964±0.0033} & 0.2822±0.0024       & {\ul 0.2010±0.0025} & 0.2114±0.0014       & 0.4178±0.0015   \\ \hline
			\end{tabular}%
		}
		\footnotesize{$^*$ Best results are shown in boldface, while the second best results are underlined.}\\
	\end{table}

	% \usepackage{graphicx}
	% \usepackage[normalem]{ulem}
	\begin{table}[htbp]
		\caption{RMSE performance (mean±std ) of our proposed CGAIN approach on 10\% missing data.\strut}
		\label{tab:10 perc}
		\resizebox{\textwidth}{!}{%
			\begin{tabular}{llllll}
				\hline
				\textbf{Dataset}        & \textbf{CGAIN}         & \textbf{GAIN}       & \textbf{MICE}       & \textbf{MissForest} & \textbf{Matrix} \\ \hline
				\textbf{Breast Cancer}  & \textbf{0.0628±0.0024} & 0.0931±0.0010       & 0.0881±0.0054       & {\ul 0.0692±0.0017} & 0.6895±0.0038   \\
				\textbf{Spambase}       & \textbf{0.0664±0.0017} & {\ul 0.0702±0.0031} & 0.0793±0.0040       & 0.0783±0.0029       & 0.0906±0.0011   \\
				\textbf{Letter}         & \textbf{0.1057±0.0014} & 0.1309±0.0008       & 0.1878±0.0010       & {\ul 0.1103±0.0021} & 0.4539±0.0018   \\
				\textbf{Default Credit} & \textbf{0.2009±0.0022} & {\ul 0.2109±0.0344} & 0.2491±0.0085       & 0.2439±0.0079       & 0.2559±0.0075   \\
				\textbf{News}           & \textbf{0.1937±0.0074} & 0.2680±0.0015       & {\ul 0.2124±0.0013} & 0.2442±0.0015       & 0.4175±0.0016   \\ \hline
			\end{tabular}%
		}
		\footnotesize{$^*$ Best results are shown in boldface, while the second best results are underlined.}\\
	\end{table}

	% \usepackage{graphicx}
	% \usepackage[normalem]{ulem}
	\begin{table}[htbp]
		\caption{RMSE performance (mean±std ) of our proposed CGAIN approach on 15\% missing data.\strut}
		\label{tab:15 perc}
		\resizebox{\textwidth}{!}{%
			\begin{tabular}{llllll}
				\hline
				\textbf{Dataset}        & \textbf{CGAIN}         & \textbf{GAIN}       & \textbf{MICE}       & \textbf{MissForest} & \textbf{Matrix} \\ \hline
				\textbf{Breast Cancer}  & \textbf{0.0673+0.0039} & 0.0986±0.0033       & 0.0877±0.0056       & {\ul 0.0689±0.0058} & 0.7042±0.0016   \\
				\textbf{Spambase}       & \textbf{0.0607±0.0033} & {\ul 0.0739±0.0025} & 0.0784±0.0024       & 0.0777±0.0021       & 0.0902±0.0052   \\
				\textbf{Letter}         & \textbf{0.1021±0.0010} & 0.1326±0.0091       & 0.1836±0.0010       & {\ul 0.1125±0.0012} & 0.4537±0.0024   \\
				\textbf{Default Credit} & \textbf{0.2314±0.0035} & {\ul 0.2442±0.0089} & 0.2479±0.0074       & 0.2672±0.0025       & 0.2565±0.0059   \\
				\textbf{News}           & \textbf{0.1992±0.0069} & 0.2869±0.0036       & {\ul 0.2283±0.0015} & 0.2918±0.0015       & 0.4177±0.0015   \\ \hline
			\end{tabular}%
		}
		\footnotesize{$^*$ Best results are shown in boldface, while the second best results are underlined.}\\
	\end{table}

	% \usepackage{graphicx}
	% \usepackage[normalem]{ulem}
	\begin{table}[htbp]
		\caption{RMSE performance (mean±std ) of our proposed CGAIN approach on 20\% missing data.\strut}
		\label{tab:20 perc}
		\resizebox{\textwidth}{!}{%
			\begin{tabular}{llllll}
				\hline
				\textbf{Dataset}        & \textbf{CGAIN}         & \textbf{GAIN}       & \textbf{MICE}       & \textbf{MissForest} & \textbf{Matrix} \\ \hline
				\textbf{Breast Cancer}  & \textbf{0.0637±0.0092} & 0.1053±0.0046       & 0.0903±0.0064       & {\ul 0.0726±0.0038} & 0.6858±0.0012   \\
				\textbf{Spambase}       & \textbf{0.0601±0.0013} & {\ul 0.0764±0.0034} & 0.0796±0.0032       & 0.0786±0.0059       & 0.0896±0.0019   \\
				\textbf{Letter}         & \textbf{0.1040±0.0027} & 0.1302±0.0031       & 0.1886±0.0010       & {\ul 0.1163±0.0013} & 0.4545±0.0028   \\
				\textbf{Default Credit} & \textbf{0.2213±0.0099} & {\ul 0.2426±0.0090} & 0.2480±0.0091       & 0.2646±0.0026       & 0.2537±0.0051   \\
				\textbf{News}           & \textbf{0.1931±0.0014} & 0.2686±0.0010        & {\ul 0.2424±0.0022} & 0.3907±0.0015       & 0.4176±0.0015   \\ \hline
			\end{tabular}%
		}
		\footnotesize{$^*$ Best results are shown in boldface, while the second best results are underlined.}\\
	\end{table}
	
	\subsection{Performance of CGAIN on imbalanced data}
	We discussed in Section \ref{sec: related work} that our proposed CGAIN approach takes into account the individual class distributions to impute the missing data. Therefore, we expect CGAIN to improve the RMSE of individual classes present in the data. We performed this experiment on the \emph{Default Credit} dataset and present the results in Table \ref{tab:imbalanced}. For this experiment, we randomly deleted rows of data belonging to a class, to introduce imbalance in the data. We tested the imputation performance using 10\%, 25\%, 40\%, and 50\% of data belonging to the minority class (see column 1 of Table \ref{tab:imbalanced}, where $n_0$ and $n_1$ show the number of samples belonging to the majority  and minority class, respectively). As with the previous experiments, we deleted 20\% of data (in a MCAR manner) to induce missingness. We also validated this experiment on unseen test data using 10-fold cross validation. 
	
	Table \ref{tab:imbalanced} shows that our proposed CGAIN approach consistently outperforms the GAIN and other imputation approaches. As the data becomes more imbalanced (e.g. $n_1=10\%$), our proposed CGAIN approach provides superior RMSE of 0.2462±0.0057 compared with 0.2632 ± 0.0076, 0.2685 ± 0.0047, 0.2934 ± 0.0095, and 0.2934 ± 0.0095 for GAIN, MICE, MissForest, and Matrix completion approach, respectively. Table \ref{tab:imbalanced} shows that the improvement in RMSE of our approach arises from the improvement in the RMSE of the individual classes present in the data. As such, our proposed CGAIN approach provides better imputation performance for balanced as well as imbalanced datasets.
	
	% \usepackage{multirow}
	% \usepackage{graphicx}
	\begin{table}[h]
		\caption{Comparative performance {[}RMSE (mean±std){]} of our proposed CGAIN approach on various imbalanced versions of the Default Credit dataset.\strut}
		\label{tab:imbalanced}
		\resizebox{\textwidth}{!}{%
			\begin{tabular}{lllllll}
				\hline
				\textbf{\begin{tabular}[c]{@{}l@{}}Proportion of classes\end{tabular}}          & \textbf{Class} & \textbf{CGAIN}         & \textbf{GAIN} & \textbf{MICE} & \textbf{MissForest} & \textbf{Matrix} \\ \hline
				\multirow{2}{*}{\begin{tabular}[c]{@{}l@{}}$n_0 = 90\%, n_1 = 10\%$\\ \end{tabular}} & Class 0 & \textbf{0.2197±0.0022} & 0.2384±0.0034 & {\ul 0.2203±0.0070} & 0.2304±0.0076       & 0.2317±0.0066   \\
				& Class 1 & \textbf{0.2462±0.0057} & {\ul 0.2632±0.0076} & 0.2685±0.0047 & 0.2934±0.0095       & 0.2847±0.0072   \\
				\multirow{2}{*}{\begin{tabular}[c]{@{}l@{}}$n_0 = 75\%, n_1 = 25\%$ \end{tabular}} & Class 0 & \textbf{0.2204±0.0031} & 0.2385±0.0014 & {\ul 0.2280±0.0055} & 0.2320±0.0059       & 0.2327±0.0022   \\
				& Class 1 & \textbf{0.2319±0.0054} & {\ul 0.2491±0.0056} & 0.2580±0.0031 & 0.2760±0.0051       & 0.2689±0.0073   \\
				\multirow{2}{*}{\begin{tabular}[c]{@{}l@{}}$n_0 = 60\%, n_1 = 40\%$\end{tabular}} & Class 0 & \textbf{0.2224±0.0024} & 0.2450±0.0019 & {\ul 0.2321±0.0086} & 0.2391±0.0042       & 0.2384±0.0083   \\
				& Class 1 & \textbf{0.2174±0.0097} & {\ul 0.2301±0.0025} & 0.2498±0.0072 & 0.2538±0.0076       & 0.2580±0.0063   \\
				\multirow{2}{*}{\begin{tabular}[c]{@{}l@{}}$n_0 = 50\%, n_1 = 50\%$\end{tabular}} & Class 0 & \textbf{0.2300±0.0027} & 0.2514±0.0009 & {\ul 0.2353±0.0016} & 0.2450±0.0044       & 0.2458±0.0028   \\
				& Class 1 & \textbf{0.1963±0.0077} & {\ul 0.2238±0.0012} & 0.2412±0.0012 & 0.2459±0.0076       & 0.2577±0.0051   \\ \hline
			\end{tabular}%
		}
		\footnotesize{$^*$ Best results are shown in boldface, while the second best results are underlined.}\\
	\end{table}
	
	\subsection{Computational cost analysis}
\begin{figure}[htbp]
		\centerline{\includegraphics[width=80mm]{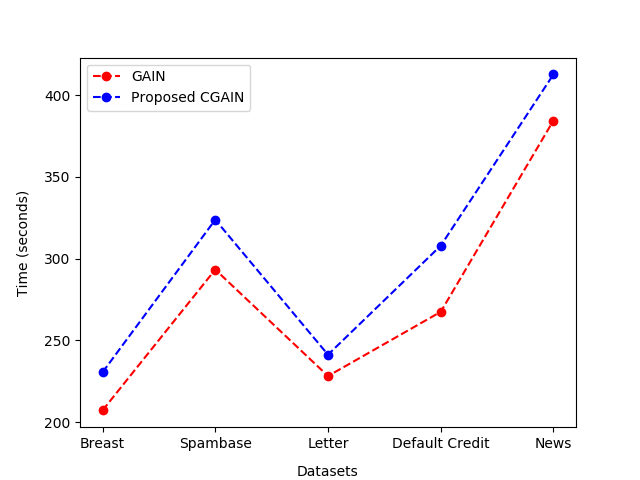}}
		\caption{Computational time analysis of GAIN and our proposed CGAIN approach.}
		\label{fig: time}
	\end{figure}

	A comparison of the computational time taken by the state-of-the-art GAIN approach and our proposed CGAIN approach is shown in Figure \ref{fig: time}. We performed all our experiments on a core i7 machine supported by an NVIDIA Quadro P5000 Graphics Processing Unit (GPU). Figure \ref{fig: time} shows the total time taken to perform 10 fold cross-validation of a dataset using GAIN or CGAIN approach. Our approach takes slightly more time compared with the GAIN approach, which is reasonable considering the use of label encoding in both the generator and discriminator in our approach.
	
	\section{Conclusion}
	\label{sec: conclusion}
	%The state-of-the-art generative imputation approaches model the distribution of complete data (as a whole) which does not take into account the outcome/class-specific distributions. The class-specific distributions contain useful information about the classes which can be helpful in approximating the missing values. This is especially useful in the scenarios where the data is imbalanced (most of the samples belong to one class and few samples to other classes). 
	In this work, we have proposed a CGAIN approach which conditions the missing data imputation on class labels using label encoding. This allows our approach to learn class-specific distributions to impute the missing values especially in imbalanced scenarios. Our CGAIN approach shows superior imputation performance compared with popular approaches on publicly available datasets.
	
	\section*{Acknowledgment}
	
	This work is partially supported by Australian Research Council Grants DP150100294 and DP150104251, and the UWA SIRF scholarship. We thank the contributors of UCI machine learning repository who collected the data and made it publicly available. We also acknowledge the computing support provided as a Quadro P5000 GPU by the NVIDIA Corporation.
	
	% \medskip
	\newpage
	\bibliographystyle{plain}
	\bibliography{references}
	\newpage

\end{document}